\documentclass{article} %
\usepackage{arxiv}

\usepackage{amsmath,amsfonts,bm}

\def\eqref#1{equation~\ref{#1}}

\def\1{\bm{1}}

\DeclareMathAlphabet{\mathsfit}{\encodingdefault}{\sfdefault}{m}{sl}
\SetMathAlphabet{\mathsfit}{bold}{\encodingdefault}{\sfdefault}{bx}{n}

\usepackage[utf8]{inputenc} %
\usepackage[T1]{fontenc}    %
\usepackage{hyperref}       %
\usepackage{url}            %
\usepackage{booktabs}       %
\usepackage{amsfonts}       %
\usepackage{microtype}      %
\usepackage{xcolor}         %
\usepackage{multirow}

\usepackage{graphicx}
\usepackage{tcolorbox}
\usepackage{wrapfig}

\definecolor{findborder}{HTML}{14403B} 
\definecolor{findbg}{HTML}{F2F2F2}     
\newtcolorbox{findingbox}{
    enhanced,                  
    colframe=findborder,       
    colback=findbg,            
    boxrule=1pt,               
    arc=4pt,                   
    drop shadow=black!30,      
    left=8pt,                  
    right=8pt,                 
    top=6pt,                   
    bottom=6pt,                
    before skip=12pt,          
    after skip=12pt            
}

\date{}

\begin{document}

\maketitleblock
  {The Hidden Evolution of Disguised Visual Context inside the VLM}
  {Wish Suharitdamrong\textsuperscript{1} \quad
   Tony Alex\textsuperscript{1,2} \quad
   Xiatian Zhu\textsuperscript{1,2} \quad
   Muhammad Awais\textsuperscript{1,2} \quad
   Sara Atito\textsuperscript{1,2,*}}
  {\{ws00372, t.alex, xiatian.zhu, muhammad.awais, sara.atito\}@surrey.ac.uk}
  {\textsuperscript{1}Surrey Institute for People-Centred AI, University of Surrey, Guildford, GU2 7XH, UK \\
   \textsuperscript{2}Centre for Vision, Speech and Signal Processing (CVSSP), University of Surrey \\
   \textsuperscript{*}Corresponding author}

\begin{abstract}
Visual tokens enter Large Language Models (LLMs) as raw, foreign signals. How they are transformed into meaningful representations and interact with the language space depends entirely on the integration architecture. Whether by treating visual tokens as in-context prompts within the input sequence or injecting them directly into the LLM's intermediate layers. A controlled comparison and understanding of how these architectural choices affect visual information and its internal transformation to integrate with the LLM remains underexplored. We provide a fair comparison by evaluating in-context and layer-wise injection VLM integration paradigms under identical training conditions across single image, multi-image, and video benchmarks. In doing so, we uncover a hidden evolution where visual tokens enter the LLM as disguised visual context, raw representations lacking linguistic structure, but are progressively reshaped depending on the integration paradigm, each capturing fundamentally different frequency characteristics of the visual signal. We show that this evolution inside the LLM determines what visual features the VLM can utilize effectively, how visual representations align with the language space, and ultimately how each paradigm performs across different tasks. We further demonstrate that attention allocation alone does not explain these performance differences. Instead, performance depends on the quality of the visual representations that each paradigm shapes at every layer. We confirm this by a direct causal intervention on the representations, whose effect differs sharply across paradigms.
\end{abstract}

\section{Introduction}
\label{sec:intro}

Vision-Language Models (VLMs) ~\citep{bai2025qwen3,deitke2025molmo}  succeed only when an LLM can reliably use visual information during generation; critically, integration design determines what ``use'' even means inside the model. 
The common approach concatenates visual tokens with text at the input so that visual tokens participate in standard self-attention throughout the network \citep{li2023blip,merullo2022linearly,eichenberg2022magma,tsimpoukelli2021multimodal,manas2023mapl}; we call this \textbf{in-context injection}.
A competing family introduces visual information at intermediate layers through dedicated mechanisms \citep{alayrac2022flamingo,li2025otter}; we call this \textbf{layer-wise injection}.
Orthogonal to this integration choice, modern frameworks \citep{liu2023visual,liu2024improved} increasingly unfreeze both the vision encoder and the LLM backbone for end-to-end training, driving rapid progress across image understanding~\citep{tong2024cambrian,wang2024cogvlm,lin2023sphinx}, video understanding~\citep{lin2024video}, and unified frameworks~\citep{li2024llava,zhang2025videollama}. As VLMs move toward long-context and text-centric settings such as OCR documents and video, integration decisions increasingly control both computational cost and what visual evidence can be composed and retained.

Despite the crucial role of integration design, we still lack controlled evidence about what the integration paradigm changes mechanistically inside the LLM. 
Most existing comparisons ~\citep{deitke2025molmo,tong2024cambrian, an2025llava} are confounded by simultaneous differences in training data mixtures, token budgets, model scale, and optimization, making it difficult to attribute performance differences to the integration mechanism itself. In addition, much of the analysis literature ~\citep{shukor2024implicit,chen2024image,basu2024understanding, neo2024towards}  focuses on a single dominant paradigm, leaving unclear whether conclusions about representation evolution, modality alignment, or attention-based usage generalize to alternative integration strategies. As a result, integration choices are often made based on convention or efficiency considerations without a clear understanding of the representational tradeoffs they induce.

This paper isolates integration as the key variable by evaluating representative in-context and layer-wise injection architectures under identical training conditions across single-image, multi-image, and video benchmarks. Beyond reporting benchmark outcomes, we provide a mechanistic characterization of how visual information is processed inside the LLM through five analyses:
\begin{itemize}
    \item \textbf{Representation evolution:} we test whether visual token representations evolve smoothly across layers.
    \item \textbf{Feature content across depth:} we examine what visual features are captured across depth via frequency-based analysis.
    \item \textbf{Convergence to language space:} we study whether visual representations converge toward the language space.
    \item \textbf{Visual information usage during generation:} we analyze when the model utilizes visual information during generation and show that attention allocation alone is insufficient to explain performance gaps.
    \item \textbf{High-frequency suppression:} we suppress the high-frequency content of visual representations, confirming that these features drive in-context injection's advantage on text-centric tasks.
\end{itemize}

These results show that integration is not merely a question of where visual features enter the network, but a choice that determines how visual information is represented and made accessible to language generation. This mechanistic perspective, causally validated by our frequency intervention, helps explain why integration strategies exhibit distinct strengths across tasks, especially in settings that require composing fine-grained evidence across many visual tokens, such as OCR and video. Our analysis provides a principled basis for selecting integration paradigms and designing future VLM architectures beyond benchmark-driven trial and error.

\section{Background}

\subsection{Integration in MLLMs}
The architecture of Multimodal Large Language Models (MLLMs) is largely defined by how modality tokens are delivered to the LLM backbone. Existing approaches can be broadly categorized into three paradigms. In-context injection~\citep{liu2024improved,zhang2025videollama} concatenates modality tokens with text tokens at the LLM input layer. Layer-wise injection introduces modality information at intermediate LLM layers through gated cross-attention~\citep{alayrac2022flamingo,laurenccon2023obelics}, adaptive gating~\citep{ye2024mplug}, LayerNorm injection~\citep{yue2025lavi}, or direct injection into the LLM's attention~\citep{alex2025pal,zhang2025layer}. Hybrid approaches~\citep{hong2024cogagent,alex2025pal} feed compressed modality tokens as an in-context prompt while injecting finer-grained representations at intermediate layers. Orthogonal to these integration paradigms, other works explore visual feature enhancement, injecting features from different levels of the vision encoder into different LLM layers~\citep{meng2024deepstack,lin2025multi}, or bypass the external vision encoder entirely through encoder-free approaches~\citep{diao2025evev2,luo2025mono,wang2025vision}, processing raw pixels directly in the LLM. In this study, we focus on in-context and layer-wise injection within the vision-language setting, the two most widely employed integration strategies.

\subsection{Previous Work in MLLM Analysis}
A growing body of work analyzes how visual information is processed inside the LLM of MLLMs, revealing an implicit multimodal alignment despite a persistent modality gap~\citep{shukor2024implicit}, visual representations becoming increasingly interpretable across layers via the logit lens~\citep{neo2024towards,jiang2024interpreting,jiang2025devils}, and middle layers critical for cross-modal interaction~\citep{kaduri2025s,jiang2025devils,yu2025multimodal,basu2024understanding}. Further studies show that visual tokens are primarily utilized in shallow layers, diminishing with depth~\citep{chen2024image}, and exhibit a visual attention sink~\citep{kang2025see}, while others map the cross-modal information flow~\citep{zhang2025cross,kim2026map}. These works predominantly focus on a single integration paradigm, typically in-context injection. Closest to our study, \citet{shukor2023ep} freeze most parameters and train only a linear projection with a single prepended token, \citet{vallaeys2024improved} train only lightweight adaptation modules with a frozen LLM and encoder, and \citet{laurenccon2024matters} rely on LoRA fine-tuning. In contrast, our comparison performs full end-to-end fine-tuning of the vision encoder, projector, and LLM, reflecting modern VLM training practice. While these works compare integration mechanisms at the benchmark level, a systematic comparison of how different integration strategies affect visual information processing within the LLM remains unexplored.

\section{Model Architectures and Training Data}

\subsection{Integration Paradigms and Architectures}

\begin{figure}[h]
  \centering
  \includegraphics[width=0.7\linewidth]{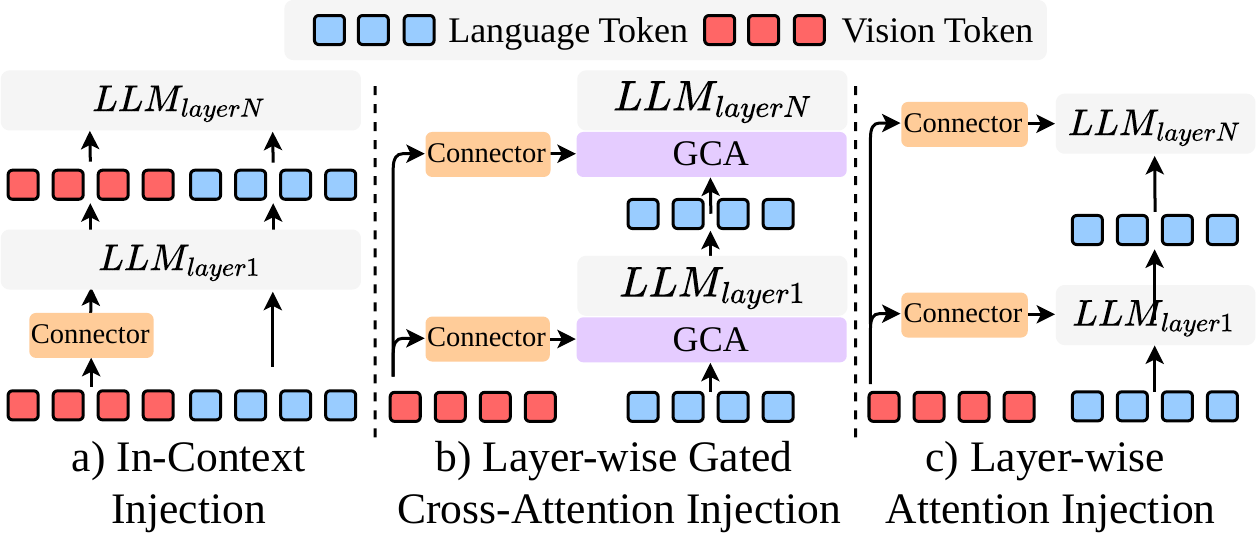}
  \caption{Overview of the three VLM integration paradigms evaluated in this study. (a) IN-CT concatenates visual tokens with text tokens at the input layer, (b) LW-GC introduces visual features through gated cross-attention (GCA) blocks, and (c) LW-AT injects visual features directly into the LLM's keys and values.}
  \label{fig:fig1_vlm_integration}
\end{figure}

To conduct a fair systematic evaluation, we standardize the training data and optimization recipe across all models, so that observed differences can be attributed primarily to the integration design rather than confounding factors. 
We focus on integration paradigms that are both widely adopted and representative of distinct design philosophies for delivering visual context into an LLM. Concretely, we employ three distinct baselines representing the primary vision-language integration paradigms as illustrated in Figure~\ref{fig:fig1_vlm_integration}.
First, in-context injection which treats visual tokens as part of the LLM input sequence. We represent this category with \textbf{IN-CT} \citep{liu2024improved}, which concatenates projected visual tokens with the text tokens.
Second, gated cross-attention layer-wise injection  which introduces visual information through dedicated cross-attention blocks inserted between the LLM layers. We represent this category with \textbf{LW-GC} \citep{alayrac2022flamingo}.
Third, attention-only layer-wise injection which incorporates visual information directly into the LLM's existing self-attention mechanism without introducing separate cross-attention blocks. We represent this category with \textbf{LW-AT} \citep{alex2025pal}. 
Together, these three architectures capture the primary integration choices found in the vision-language literature.

Given an input image $V$, a vision encoder $\Phi$ extracts a sequence of $N_v$ visual features, $Z_v = \Phi(V) \in \mathbb{R}^{N_v \times d_v}$, where $d_v$ is the output dimension of the vision encoder. All three paradigms share the same connector architecture, a simple two-layer MLP with GELU activation: $ \tilde{Z}_v = \Psi(Z_v) = W_{\mathrm{out}}\operatorname{GELU}(W_\mathrm{in} \cdot Z_v)$, where $\tilde{Z}_v \in \mathbb{R}^{N_v \times d_t}$ denotes the projected visual tokens, $d_t$ is the hidden dimension of the LLM, and $W_\mathrm{in}$ and $W_\mathrm{out}$ are learnable projection matrices. Bias terms are omitted for clarity. For LW-GC, we do not utilize a Perceiver~\citep{jaegle2021perceiver} nor compress the number of visual tokens, maintaining an identical number of visual tokens across all paradigms. The three integration mechanisms are detailed as follows:

\begin{enumerate}

\item \textbf{IN-CT (In-context injection):} A single connector projects the visual features into the LLM embedding space, producing $\tilde{Z}_v$. Let $H_t \in \mathbb{R}^{N_t \times d_t}$ denote the embeddings of the $N_t$ text tokens. The initial hidden state is formed by concatenating the projected visual and text tokens, $h_0 = [\tilde{Z}_v; H_t]$, after which the combined sequence is processed jointly through the LLM's standard transformer blocks, as shown in Figure~\ref{fig:fig1_vlm_integration}(a).
    
\item \textbf{LW-GC (Layer-wise with gated cross-attention):} The initial hidden-state sequence contains only the text tokens embeddings ($h_0 = H_t$). At each layer $l$, a layer-specific connector $\Psi_l$ maps visual features, which are injected via a gated cross-attention block (CA) and a gated feed-forward network (FFN), both modulated by learnable scalars $\alpha_l$ and $\beta_l$ initialized to zero:
\begin{align}
  h^{\text{cross}}_{l} &= h_l + \tanh(\alpha_l)\cdot \text{CA}(h_l,\,  \tilde{Z}_v^l,\,  \tilde{Z}_v^l), \label{eq:lwgc_cross}\\
  h^{\text{gated}}_{l} &= h^{\text{cross}}_{l} + \tanh(\beta_l)\cdot \text{FFN}(h^{\text{cross}}_{l}). \label{eq:lwgc_gated}
\end{align}
We evaluate two injection schedules: gated cross-attention blocks inserted at every LLM layer and blocks inserted once every four layers. Refer to Figure~\ref{fig:fig1_vlm_integration}(b).

\item \textbf{LW-AT (Layer-wise with attention-only injection):} Starting with $h_0 = H_t$, layer-specific connectors map visual features that are injected solely into the LLM's existing attention, explicitly bypassing the FFNs:
\begin{equation}
  h'_{l} = \text{Attn}\big(h_l,\, [\tilde{Z}_v^l; h_l],\, [\tilde{Z}_v^l; h_l]\big). \label{eq:lwat}
\end{equation}
LW-AT bridges the two prior paradigms: like LW-GC, visual features are injected layer-by-layer, but like IN-CT, visual and text tokens interact simultaneously within standard attention as shown in Figure~\ref{fig:fig1_vlm_integration}(c).
\end{enumerate}

Further implementation details and training settings for all three paradigms are provided in Appendix ~\ref{app:a1_implementation_details} and ~\ref{app:a2_training_details}.

\subsection{Model and Dataset}
We use Qwen2.5-Instruct-0.5B, 3B, and 7B~\citep{qwen2.5} and LLaMA-3.2-Instruct-1B and 3B ~\citep{grattafiori2024llama} as LLM backbones and SigLIP2-So400m~\citep{tschannen2025siglip} as the vision encoder. Training proceeds in two stages: (i) connector pretraining and (ii) supervised fine-tuning (instruction tuning). For connector pretraining, we use the BLIP-558K dataset~\citep{liu2023visual}. For instruction tuning, we employ two data mixtures: (1) the \textbf{LLaVA-NeXT} instruction tuning data~\citep{liu2024llavanext} comprising $\sim$700K samples, and (2) the \textbf{LLaVA-OneVision (OV)} instruction tuning data~\citep{li2024llava}, a comprehensive mixture of single images, multi-image, and videos comprising $\sim$1.6M samples. Throughout the paper, we refer to these two instruction tuning mixtures as \textbf{LLaVA-NeXT} and \textbf{LLaVA-OV}, respectively.

\section{Benchmark Results}
\label{sec:section2_benchmark_results}

\begin{figure*}[h]
  \centering
  \includegraphics[width=\linewidth]{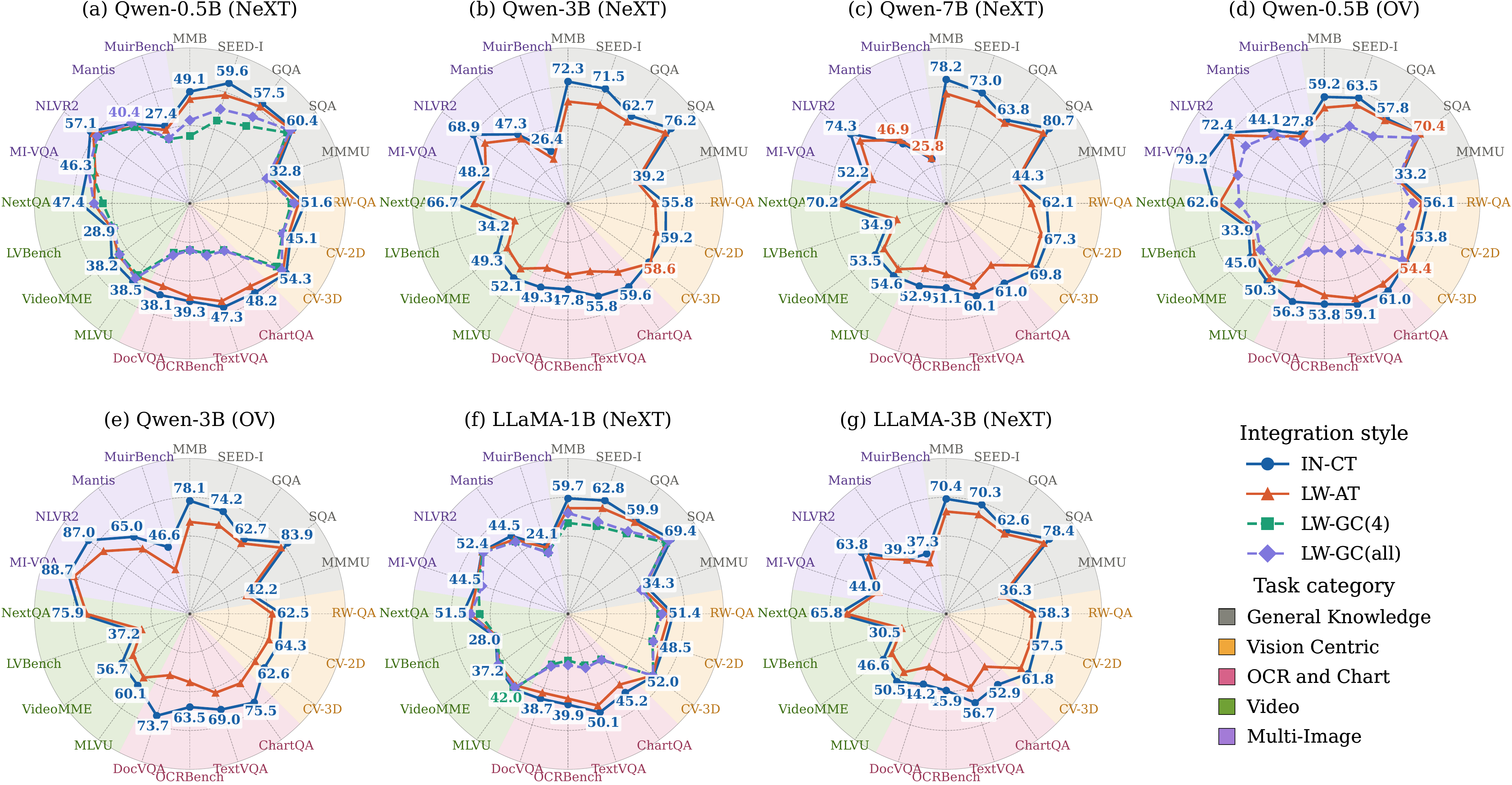}
  \caption{Benchmark comparison of the three integration paradigms across single-image, multi-image, and video understanding. Panels correspond to backbone and training setup: Qwen2.5-Instruct (0.5B, 3B, 7B) and LLaMA-3.2-Instruct (1B, 3B) with LLaVA-NeXT, and Qwen2.5-Instruct with LLaVA-OV on 0.5B and 3B. Axes are grouped by task type, and the annotated value on each axis is the best score on that benchmark within the panel.}
  \label{fig:fig2_main_radar}
\end{figure*}

We evaluate the three integration paradigms on benchmarks spanning single-image understanding (general knowledge, vision-centric, and OCR), multi-image understanding, and video understanding. As shown in Figure~\ref{fig:fig2_main_radar}, IN-CT achieves the strongest overall performance, with LW-AT following closely and occasionally surpassing it on individual tasks. LW-GC, in contrast, lags behind both by a wide margin. This trend holds across both LLM backbones, all model scales, and both instruction tuning mixtures (LLaVA-NeXT and LLaVA-OV), indicating that the performance differences stem from the integration paradigm rather than from any specific LLM backbone, size, or training data. 

The gap between IN-CT and the layer-wise variants is largest specifically on the OCR and video tasks, while performance remains close elsewhere. Since LW-AT shares the same token interaction mechanism as IN-CT, differing only in how visual tokens are injected and passed to the LLM, this gap points directly at the injection choice. The gap in video tasks can be explained by the need to integrate information across frames. Inside the LLM, attention among visual tokens across frames is causally necessary for capturing temporal information~\citep{kim2026map}, a capability that in-context injection supports through joint self-attention but that layer-wise injection cannot provide, since its visual tokens remain static and do not attend to one another. On OCR, this gap is even more striking. For OCR, both layer-wise variants consistently fall behind IN-CT by a large margin, especially LW-GC.

\subsection{Effect of Text-Centric Data}

\begin{wrapfigure}{r}{0.5\textwidth}
  \vspace{-\intextsep}
  \centering
  \includegraphics[width=0.48\textwidth]{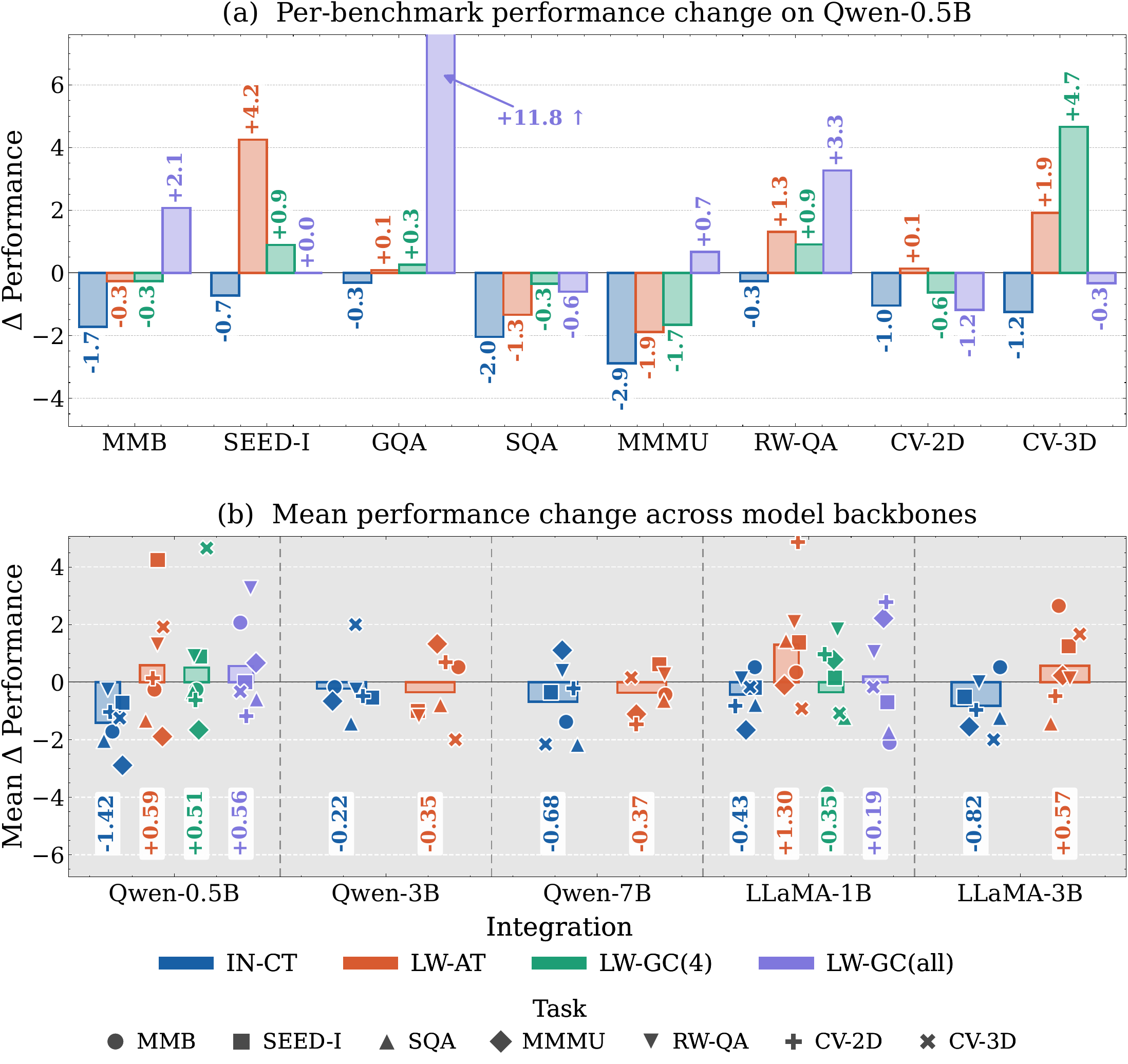}
  \caption{Effect of removing text-centric data. (a) Per-benchmark change
  ($\Delta$ score) of the Qwen 0.5B model trained on the text-centric-filtered
  mixture relative to the full LLaVA-NeXT mixture. (b) Mean of these $\Delta$
  scores per backbone, with bars showing the mean per integration method and
  dots the individual benchmarks.}
  \label{fig:fig3_ocr_diff}
  \vspace{-\intextsep}
\end{wrapfigure}

Based on these observations, we hypothesize that text-centric visual data may provide limited benefit to the optimization of layer-wise injection models. To test this, we remove text-centric training data from the LLaVA-NeXT mixture (the filtered datasets are listed in Appendix~\ref{app:a2_datasets}) and compare the resulting performance change. Figure~\ref{fig:fig3_ocr_diff} reveal a clear difference in how each architecture responds. For IN-CT, removing these samples leads to consistent performance decreases across nearly every benchmark, whereas for layer-wise variants the impact is more selective, with both variants even improving on several benchmarks. This suggests that in-context injection can broadly leverage text-centric data to benefit general visual understanding, whereas layer-wise injection models do not utilize such data effectively, and on some tasks the data may even interfere with their optimization. We attribute this advantage to in-context visual tokens participating in the LLM's attention layers, allowing tokens from different image patches to attend to one another and progressively integrate spatially distributed information, such as characters spanning multiple patches, into coherent representations. Layer-wise injection cannot replicate this behavior, as its visual tokens never attend to one another.


\section{The Hidden Evolution of Visual Context}
The experimental results reveal a consistent performance gap between in-context injection and layer-wise injection, particularly in OCR and video tasks. What causes this gap? To understand this, we conduct five analyses examining (1) whether visual tokens evolve inside the LLM, (2) what features they capture, (3) whether they align with the language space, (4) when the model utilizes visual information, and (5) whether higher-frequency content causally drives performance. Unless stated otherwise, all analyses use the Qwen2.5-Instruct-0.5B backbone trained on the LLaVA-NeXT mixture, with LW-GC(all) as the gated cross-attention variant, since every-layer insertion provides multimodal interaction at every LLM layer, matching IN-CT and LW-AT.

\subsection{Do Visual Tokens Evolve Inside the LLM?}\label{sec:finding1_cka}

Given that IN-CT consistently outperforms layer-wise injection, we hypothesize that visual tokens in IN-CT undergo a progressive transformation through the LLM layers, similar to how text tokens are refined. To test this, we utilize Centered Kernel Alignment (CKA) \citep{kornblith2019similarity}, which unlike cosine similarity used in prior work \citep{shukor2024implicit,zhang2025layer}, is invariant to scaling and rotation, capturing structural shifts across entire token sets rather than individual vectors. Specifically, we compute CKA between the token representations of each sample at layers $\ell$ and $\ell'$, averaged over all samples (formal definitions in Appendix ~\ref{app:cka}).
As shown in Figure~\ref{fig:Fig4_fft_analysis_and_fft_analysis}(a-b), visual tokens in IN-CT exhibit a smooth, uniform transformation across layers. Text tokens are equally smooth but evolve at different depths, shifting most in the early layers while visual representations change most in the later layers. In contrast, the visual tokens of LW-AT and LW-GC exhibit severe discontinuities across layers, even though their text tokens remain smooth. The difference lies in how visual information is delivered. In IN-CT, visual tokens reside in the residual stream, so each layer introduces only incremental changes. In layer-wise injection, visual tokens bypass the residual stream and are supplied through independent per-layer projections of the encoder features, without any constraint linking consecutive layers. While different LLM layers may require different interpretations of visual features, these results show that the two integration families arrive at fundamentally different solutions to this problem.
\begin{findingbox}
\textbf{\textit{Finding 1:}} \quad Visual tokens in IN-CT semantically evolve across LLM layers, mirroring the smooth refinement of text tokens. In contrast, layer-wise injection produces discontinuous representations across layers.
\end{findingbox}


\subsection{What Do Visual Representations Capture?}\label{sec:finding2_fft}

\begin{figure*}[t]
  \centering
  \includegraphics[width=\linewidth]{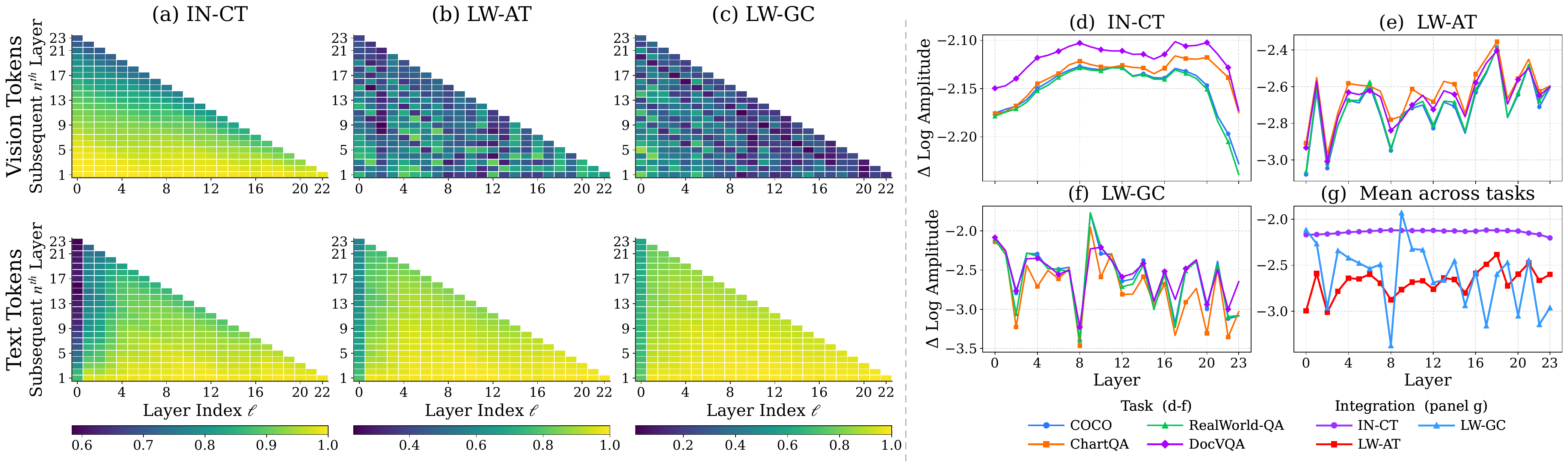}
\caption{(Left) CKA heatmaps of visual (top row) and text (bottom row) token representations across layers on ChartQA for (a) IN-CT, (b) LW-AT, and (c) LW-GC. IN-CT evolves smoothly across layers, while LW-AT and LW-GC show severe discontinuities in their visual tokens despite smooth text tokens.
(Right) Relative log amplitude of Fourier-transformed visual token representations across layers on four datasets, shown per paradigm on its own scale for (d) IN-CT, (e) LW-AT, and (f) LW-GC, and (g) averaged across tasks for all paradigms. IN-CT holds the highest amplitude, further increased for text-centric tasks, while LW-AT and LW-GC remain lower with no such task ordering.}
  \label{fig:Fig4_fft_analysis_and_fft_analysis}
\end{figure*}

Building on Finding 1, we next examine what visual features each paradigm captures across layers. Following the Fourier analysis framework \citep{park2023self}, we compute the relative log amplitude of Fourier-transformed hidden states at each layer, measuring the amplitude difference between the highest and lowest frequencies. Higher values indicate greater capture of higher-frequency information, while lower values indicate a bias toward lower-frequency information. As shown in Figure~\ref{fig:Fig4_fft_analysis_and_fft_analysis}(g), IN-CT captures and maintains higher-frequency features, holding the highest relative log amplitude across layers, whereas both layer-wise variants remain biased toward low-frequency features, with LW-AT fluctuating but relatively stable and LW-GC oscillating erratically with no coherent trend. These oscillations are consistent with the non-uniform representation shifts in Section~\ref{sec:finding1_cka}.

We next examine the per-task frequency behavior of each paradigm in Figure~\ref{fig:Fig4_fft_analysis_and_fft_analysis}(d-f). The IN-CT amplitude increases gradually through the early and middle layers, indicating a progressive shift toward fine-grained local details such as texture and edges. Moreover, its curves are consistently ordered by task type, with text-centric tasks such as DocVQA and ChartQA maintaining higher amplitudes than general vision tasks such as COCO and RealWorld-QA at nearly every layer, showing that IN-CT builds higher-frequency content for the inputs that demand fine-grained detail. The layer-wise variants lack this property, as their text-centric curves remain at similar amplitudes to the general vision tasks. Combined with their overall lower amplitudes observed above, this may explain why IN-CT's advantage is largest on OCR, where fine-grained detail matters most. Notably, the IN-CT amplitude drops sharply in the final layers, shifting back toward lower-frequency features. This aligns with Section~\ref{sec:finding1_cka}, where visual representations change most in the later layers, and suggests that the LLM consolidates fine-grained details into more abstract, global representations, potentially converging toward the language space, which we investigate in Section~\ref{sec:finding3_pca}.

\begin{findingbox}
\textbf{\textit{Finding 2:}} \quad IN-CT captures higher-frequency features, especially for tasks requiring fine-grained detail, and consolidates them into lower-frequency form in the final layers. LW-AT and LW-GC remain lower-frequency biased and show no such task differentiation.
\end{findingbox}

\subsection{Do Visual Tokens Align with the Language Representation Space?}\label{sec:finding3_pca}

\begin{figure*}[h]
  \centering
  \includegraphics[width=\linewidth]{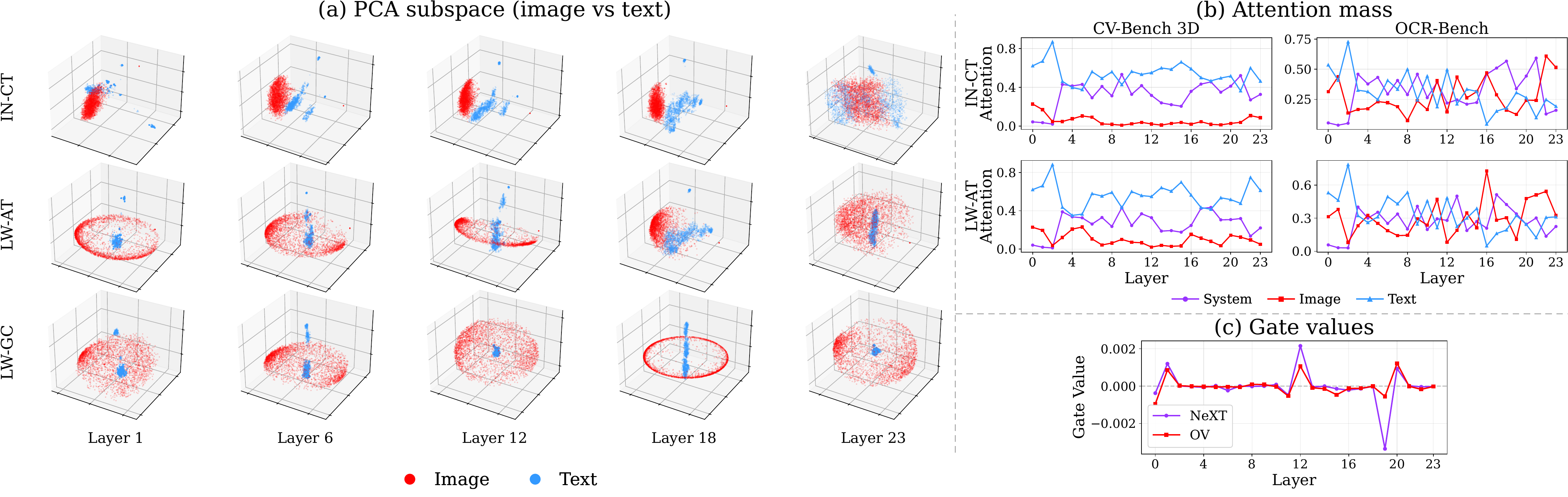}
  \caption{(Left) 3D PCA projections of image (red) and text (blue) token representations across layers for the three integration paradigms (rows: IN-CT, LW-AT, LW-GC), shown in (a). In IN-CT, the two modalities progressively merge into a shared space, whereas in LW-AT and LW-GC the image tokens remain separated from the text tokens throughout all layers. (Right) Visual information utilization. (b) Attention mass of generated tokens toward system, image, and text
tokens across layers for IN-CT and LW-AT on a general and a text-centric task. (c) Learned gate values of LW-GC
across layers under both instruction tuning mixtures.}
  \label{fig:Fig5_pca3d_plot_and_attention}
\end{figure*}

Following Section~\ref{sec:finding2_fft}, we ask whether the final-layer consolidation reflects convergence toward the language representation space. To test this, we project image and text hidden states at each layer into a shared 3D PCA space. As shown in Figure~\ref{fig:Fig5_pca3d_plot_and_attention}(a), the paradigms exhibit fundamentally different behaviors. In IN-CT, image tokens already carry rich semantic structure from the vision encoder in the early layers, while text tokens have yet to form meaningful representations. As text tokens acquire semantics in the middle layers, both modalities occupy narrow modality-specific cones, consistent with the modality gap~\citep{liang2022mind} observed in MLLMs~\citep{shukor2024implicit}. However, as tokens propagate deeper, these cones progressively merge, converging into a shared representational space by the final layers. In contrast, neither layer-wise paradigm exhibits this convergence. In both LW-AT and LW-GC, image representations remain separated from the text cluster at every depth. This is expected, as layer-wise injection is not designed to progressively transform visual tokens through the LLM, and the representations are therefore never reshaped toward the language space, consistent with the discontinuous representations observed in Section~\ref{sec:finding1_cka}.

\begin{findingbox}
\textbf{\textit{Finding 3:}} \quad Visual tokens in IN-CT progressively converge toward the language space across LLM layers, while layer-wise tokens remain separated from it.
\end{findingbox}

\subsection{When Is Visual Information Used?}\label{sec:finding4_attention}

Section~\ref{sec:finding1_cka}--\ref{sec:finding3_pca} established that IN-CT, LW-AT, and LW-GC differ fundamentally in their representations. How does the LLM utilize these representations during generation? For IN-CT and LW-AT, we quantify visual information utilization by generated tokens at each layer through Attention Mass, the total attention weight directed toward visual tokens. For a generated token at position $t$, the attention mass for a segment $S$ at layer $\ell$ is
\begin{equation}
    \text{Mass}^{(\ell)}_{S} = \frac{1}{H} \sum_{h=1}^{H} \sum_{j \in S} \alpha^{(\ell,h)}_{t,j}
\end{equation}
where $H$ is the number of attention heads and $\alpha^{(\ell,h)}_{t,j}$ is the attention weight from position $t$ to position $j$ in head $h$ of layer $\ell$.

We confirm prior findings~\citep{chen2024image,zhang2025llava} that IN-CT attends to visual tokens mainly in shallow layers, diminishing with depth, on general tasks such as CV-Bench 3D as shown in Figure~\ref{fig:Fig5_pca3d_plot_and_attention}(b), but find this pattern task-dependent. In text-centric tasks such as OCR-Bench, attention to image tokens remains high through middle and deep layers. For LW-AT, the attention distribution is surprisingly similar to IN-CT. Yet IN-CT consistently outperforms LW-AT. We attribute this gap to the quality of information at each layer rather than attention allocation. Specifically, IN-CT captures higher-frequency features (Section~\ref{sec:finding2_fft}), providing fine-grained details critical for OCR, while LW-AT remains low-frequency biased. We further hypothesize that IN-CT benefits on OCR from language-space convergence (Section~\ref{sec:finding3_pca}), as the LLM reads text-like semantics rather than raw visual features. For LW-GC, fixed learned gate values control the visual contribution regardless of input, as shown in Figure~\ref{fig:Fig5_pca3d_plot_and_attention}(c). This explains why LW-AT outperforms LW-GC despite both being low-frequency biased, as it can adapt its visual contribution through attention. We hypothesize that variants with dynamic gating~\citep{ye2024mplug} could resolve this, but would still face the same frequency limitations as LW-AT.

\begin{findingbox}
\textbf{\textit{Finding 4:}} \quad Visual information utilization by generated tokens varies across layers and is task-dependent. The performance gap cannot be explained by attention allocation alone but also reflects the quality of representations at each layer.
\end{findingbox}

\subsection{Do High-Frequency Features Provide the Benefit on Text-Centric Tasks?}
\label{sec:finding5_freq_intervention}

\begin{figure*}[!h]
  \centering
  \includegraphics[width=\linewidth]{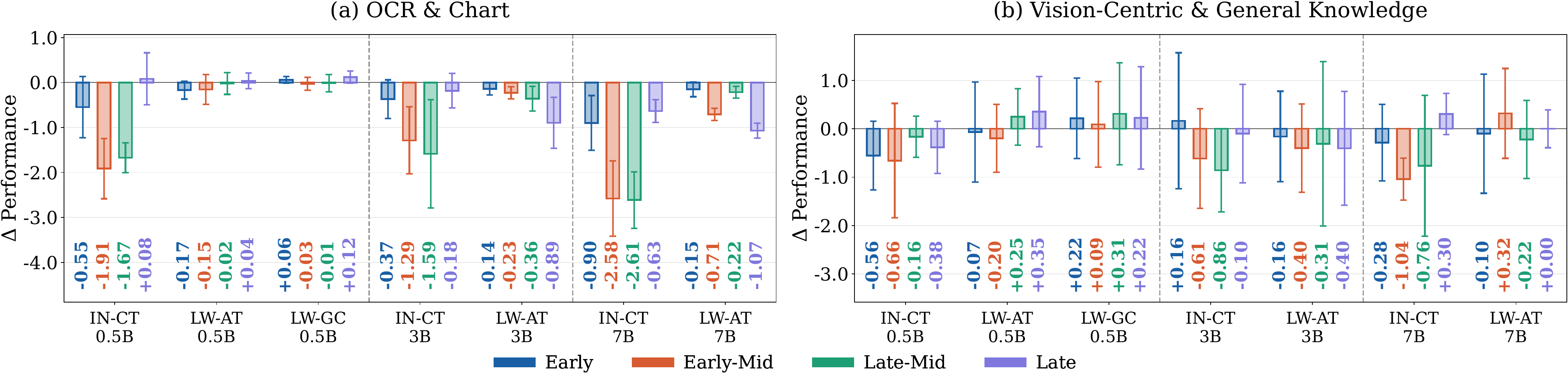}
\caption{Performance change from low-pass filtering the visual token hidden states at four decoder depth groups, shown for each model and scale on (a) OCR~\&~Chart and (b) Vision-Centric and General Knowledge tasks. Bars report the mean change relative to the unfiltered model and error bars denote variation across benchmarks within each group. IN-CT degrades most, with damage concentrated in the Early-Mid and Late-Mid groups on OCR~\&~Chart, while LW-AT is far less affected and General Knowledge remains largely unchanged.}
  \label{fig:Fig6_freq_filter_preprint}
\end{figure*}

Since IN-CT builds higher-frequency features (Section~\ref{sec:finding2_fft}) and its advantage over the layer-wise variants arises from representation quality (Section~\ref{sec:finding4_attention}), this content may itself underlie the text-centric advantage. To test this directly, we suppress the high-spatial-frequency content of visual representations inside the LLM while preserving their coarse structure and measure the downstream effect. Let $Z_\ell \in \mathbb{R}^{H \times W \times d}$ denote the visual-token hidden states at decoder layer $\ell$ on their spatial grid. We apply an ideal low-pass filter in the 2-D Fourier domain
\begin{equation}
\begin{aligned}
    \tilde{Z}_\ell &= \mathcal{F}^{-1}\big(\, M_\rho \odot \mathcal{F}(Z_\ell) \,\big), \\
    M_\rho(u,v) &=
    \begin{cases}
        1 & \text{if } \sqrt{u^2 + v^2} \le \rho \cdot k_{\max} \\
        0 & \text{otherwise}
    \end{cases}
\end{aligned}
\end{equation}
where $\mathcal{F}$ is the per-channel 2-D DFT, $(u,v)$ are the spatial frequencies measured from the spectrum center, and $k_{\max}$ is the highest representable frequency. We set the cutoff $\rho = 0.9$, removing only the highest-frequency band. To localize where fine detail matters, we filter one of four decoder depth groups (Early, Early-Mid, Late-Mid, Late) at a time and report performance change per task category, as shown in Figure~\ref{fig:Fig6_freq_filter_preprint}.

The results reveal a consistent asymmetry between the integration strategies. At every scale, the largest performance drops belong to IN-CT, while LW-AT degrades far less at matched scales and depths, and LW-GC is nearly unaffected. This asymmetry follows directly from Section~\ref{sec:finding2_fft}, as the filter suppresses precisely the content only IN-CT constructs. The degradation is most severe on OCR~\&~Chart and lesser on Vision-Centric and General Knowledge. Consistent with Section~\ref{sec:finding4_attention}, although IN-CT and LW-AT exhibit similar attention patterns, the intervention harms IN-CT far more, as the two differ in the quality of their visual representations. The tasks least reliant on fine detail show only minor degradation as shown in Figure~\ref{fig:Fig6_freq_filter_preprint}, confirming the intervention removes fine-grained information rather than broadly disrupting computation. These results thus causally confirm that the representation quality separating IN-CT from layer-wise injection is its higher-frequency content, a key source of its text-centric advantage.

\begin{findingbox}
\textbf{\textit{Finding 5:}} \quad Suppressing high-frequency visual content selectively harms IN-CT on text-centric tasks, confirming this content drives its text-centric advantage.
\end{findingbox}

\section{Ablation Study}

\begin{wrapfigure}{r}{0.5\textwidth}
  \vspace{-\intextsep}
  \centering
  \includegraphics[width=0.48\textwidth]{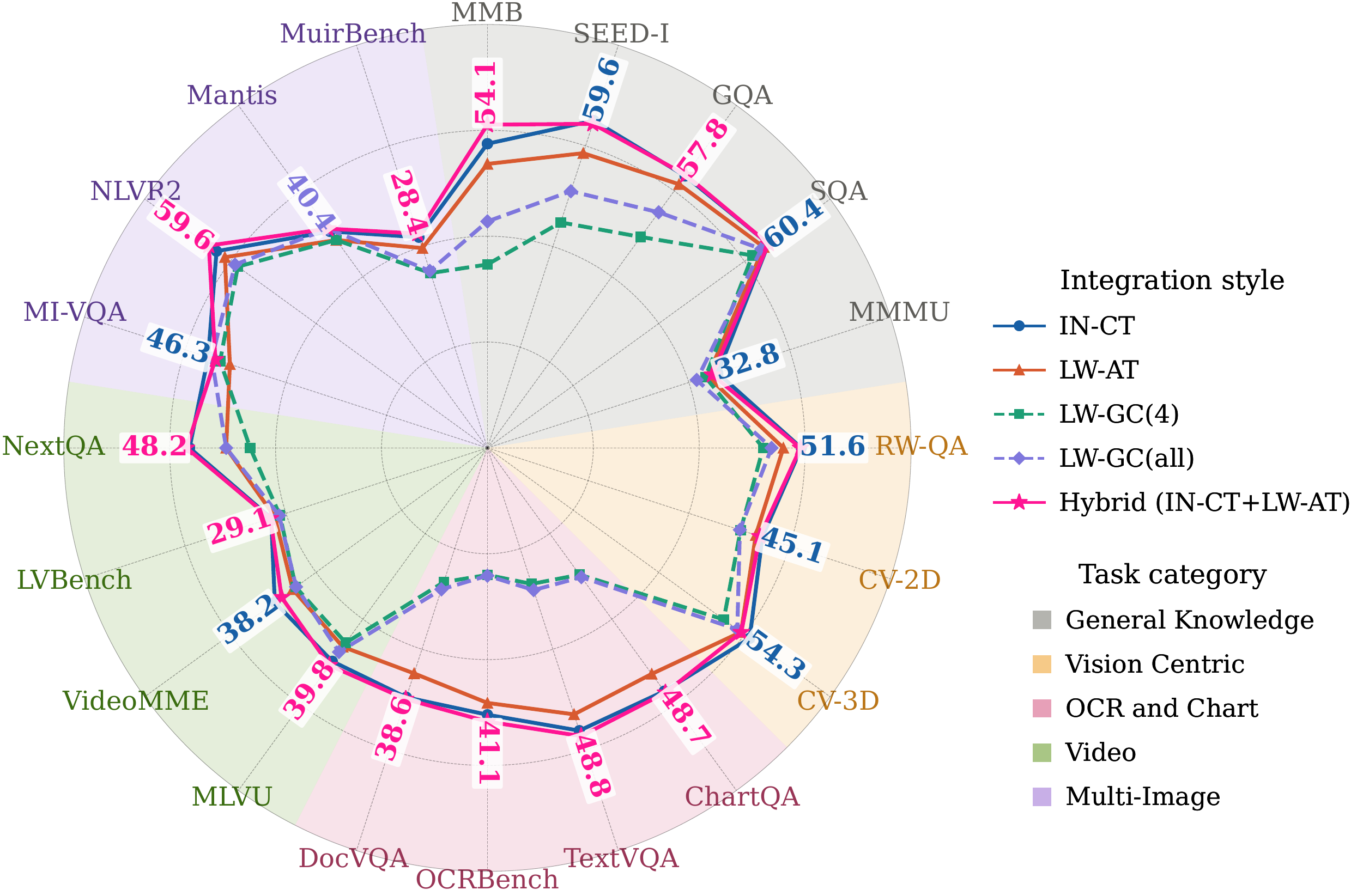}
  \caption{Benchmark performance of the hybrid IN-CT + LW-AT model compared
  with the individual integration paradigms, where the annotated value on each
  axis is the best score on that benchmark.}
  \label{fig:Fig7_hybrid_radar_preprint}
  \vspace{-\intextsep}
\end{wrapfigure}

Our analysis has shown that IN-CT and LW-AT capture different frequency characteristics, with IN-CT capturing higher-frequency while LW-AT maintains lower-frequency representations. To explore whether combining both paradigms can leverage these properties, we construct a hybrid model, IN-CT + LW-AT, applying both integration strategies simultaneously. Specifically, layer-wise attention injection is interleaved before the in-context tokens with a 1:1 token mapping (full details in Appendix ~\ref{app:a1_hybrid}). As shown in Figure~\ref{fig:Fig7_hybrid_radar_preprint}, the hybrid achieves the best performance across most benchmarks, outperforming either paradigm alone. However, it doubles the visual token count, making it computationally impractical. We present this as an exploratory finding, motivating future architectures that could combine both strategies efficiently.

\section{Conclusion}
We presented a systematic comparison of in-context and layer-wise integration paradigms in VLMs under identical training conditions. Through five analyses, we uncovered that in-context injection enables a hidden evolution of visual tokens, progressively shifting toward high-frequency features before converging with the language space, while layer-wise injection remains biased toward low-frequency features and separated from it. These representational differences, rather than attention allocation, drive the performance gap between the paradigms, as causally confirmed by our frequency intervention. Our hybrid experiment suggests a promising direction for future architectures that combine representational depth with computational efficiency. Though our study focuses on vision-language, these findings are relevant to the broader MLLM community. Different modalities may benefit from different frequency characteristics. In particular, each modality could be routed through either the efficient layer-wise pathway or the more expensive in-context pathway according to its representation's detail requirements, allowing future architectures to scale across modalities without the prohibitive bottleneck of universal in-context processing. We hope this work provides actionable insights for next-generation VLMs and MLLMs.

\noindent
\textbf{Limitations:}
All experiments use a constant-epoch budget, which does not equalize compute, as IN-CT runs visual tokens through quadratic attention and FFNs, consuming substantially more per epoch than the layer-wise variants. Under equal compute, the layer-wise methods could train longer, potentially narrowing the benchmark gaps. The representational differences we identify arise from the architecture itself and are not expected to disappear with additional training.

\section*{Acknowledgments}
This work was supported by a doctoral studentship from the Surrey Institute for
People-Centred Artificial Intelligence (PAI) at the University of Surrey. The authors acknowledge the use of resources provided by the Isambard-AI National AI Research Resource (AIRR)~\cite{mcintosh2024isambard}. Isambard-AI is operated by the University of Bristol and is funded by the UK Government’s Department for Science, Innovation and Technology (DSIT) via UK Research and Innovation; and the Science and Technology Facilities Council [ST/AIRR/I-A-I/1023].

\bibliography{iclr2026_conference}
\bibliographystyle{iclr2026_conference}

\newpage
\appendix

\section*{Appendix}

\section{Implementation Details}
\label{app:a1_implementation_details}

All three paradigms share the same vision encoder (SigLIP2-So400m~\cite{tschannen2025siglip}) and connector architecture (MLP with GELU activation), and maintain an identical number of visual tokens to ensure that any observed performance differences are attributable solely to the integration mechanism. The implementation details of each paradigm are described below.

\subsection{IN-CT}
\label{app:a1_inct}
Our IN-CT implementation directly follows~\citet{liu2024improved}. Visual and text tokens are concatenated into a single sequence of length $N_v + N_t$ and processed jointly through all LLM layers, yielding an attention complexity of $\mathcal{O}((N_v + N_t)^2)$ and an FFN cost proportional to $N_v + N_t$ at every layer. Since $N_v \gg N_t$ in practice, both costs are dominated by the visual token count.

\subsection{LW-GC}
\label{app:a1_lwgc}
Our LW-GC implementation follows~\citet{alayrac2022flamingo}, retaining the zero initialisation of the learnable gates $\alpha_l$ and $\beta_l$ in Eqs.~(\ref{eq:lwgc_cross}) and~(\ref{eq:lwgc_gated}) as in the original design. However, unlike the original Flamingo training procedure which keeps the LLM frozen, we unfreeze the LLM during supervised fine-tuning to ensure a fair comparison with IN-CT and LW-AT. During the projector pretraining stage, we train the MLP projector along with the gated cross-attention modules while keeping the LLM and vision encoder frozen. Additionally, we do not utilise a Perceiver~\citep{jaegle2021perceiver} nor compress the number of visual tokens, maintaining an identical visual token count across all three paradigms. Since the initial hidden state contains only text tokens, the standard self-attention operates over $N_t$ tokens with complexity $\mathcal{O}(N_t^2)$, while the gated cross-attention adds a cost of $\mathcal{O}(N_t \cdot N_v)$, avoiding the $\mathcal{O}(N_v^2)$ term present in IN-CT. We experiment with two insertion schedules, inserting blocks at every layer and at every fourth layer. Every-layer insertion consistently yields better performance, which we adopt as the default configuration in our main comparisons.

\subsection{LW-AT}
\label{app:a1_lwat}
\begin{figure}[h]
  \centering
  \includegraphics[width=\linewidth]{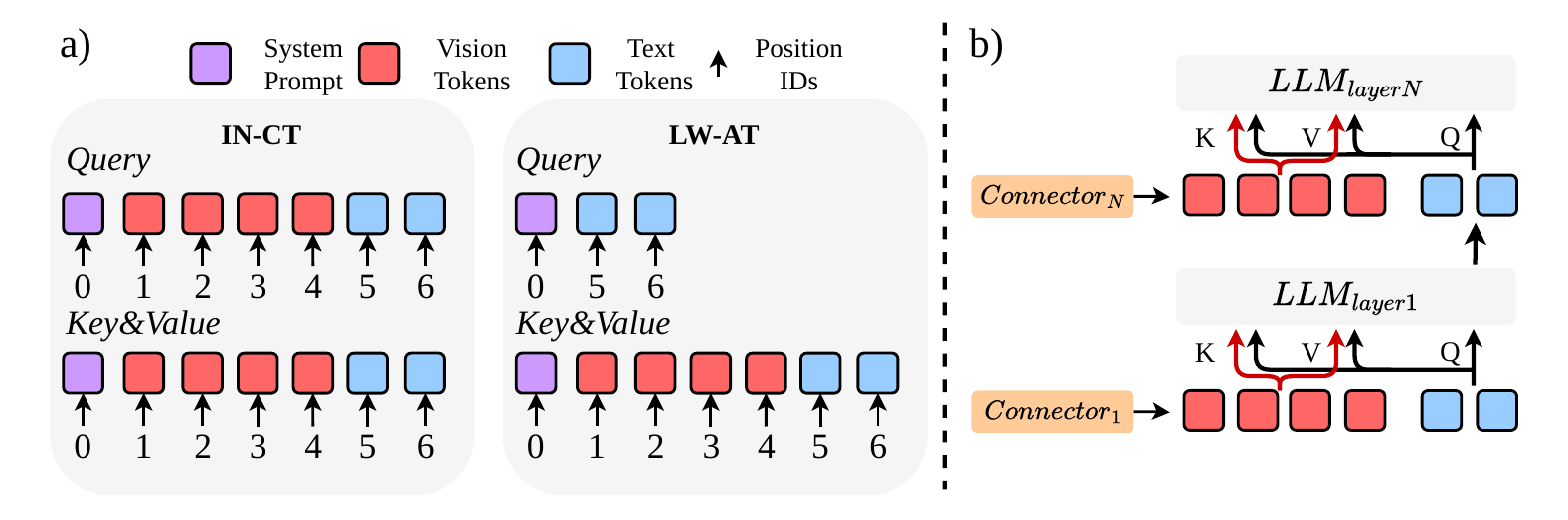}
  \caption{(a) Position ID scheme used in LW-AT compared to IN-CT. Visual tokens are assigned the same position IDs in the key-value sequence as they would occupy in IN-CT, preserving the relative positional relationship between visual and text tokens across both paradigms. (b) Architecture of LW-AT, where layer-specific connectors project visual features into the keys and values of each LLM layer, bypassing the FFN sublayers entirely.}
  \label{app_fig:appendix_lw-atn}
\end{figure}

Our LW-AT implementation, defined in Equation~(\ref{eq:lwat}), appends visual tokens to keys and values only, never forming queries, and bypasses the LLM's FFN sublayers entirely, as shown in Figure~\ref{app_fig:appendix_lw-atn}. This reduces the per-layer attention complexity from $\mathcal{O}((N_v + N_t)^2)$ in IN-CT to $\mathcal{O}((N_v + N_t) \cdot N_t)$, and since $N_v \gg N_t$ in practice, the savings are substantial. The FFN cost per layer is similarly reduced, as it operates over $N_t$ tokens rather than $N_v + N_t$. Since this is an architectural rather than a training-only modification, these efficiency gains apply at both training and inference time. We follow~\citet{alex2025pal} in adopting their position ID scheme, where visual tokens are assigned the same position IDs in the key-value sequence as they would occupy in IN-CT, as shown in Figure~\ref{app_fig:appendix_lw-atn}, ensuring a direct and fair architectural comparison between LW-AT and IN-CT.

\subsection{IN-CT + LW-AT}
\label{app:a1_hybrid}

\begin{figure}[!h]
  \centering
  \includegraphics[width=\linewidth]{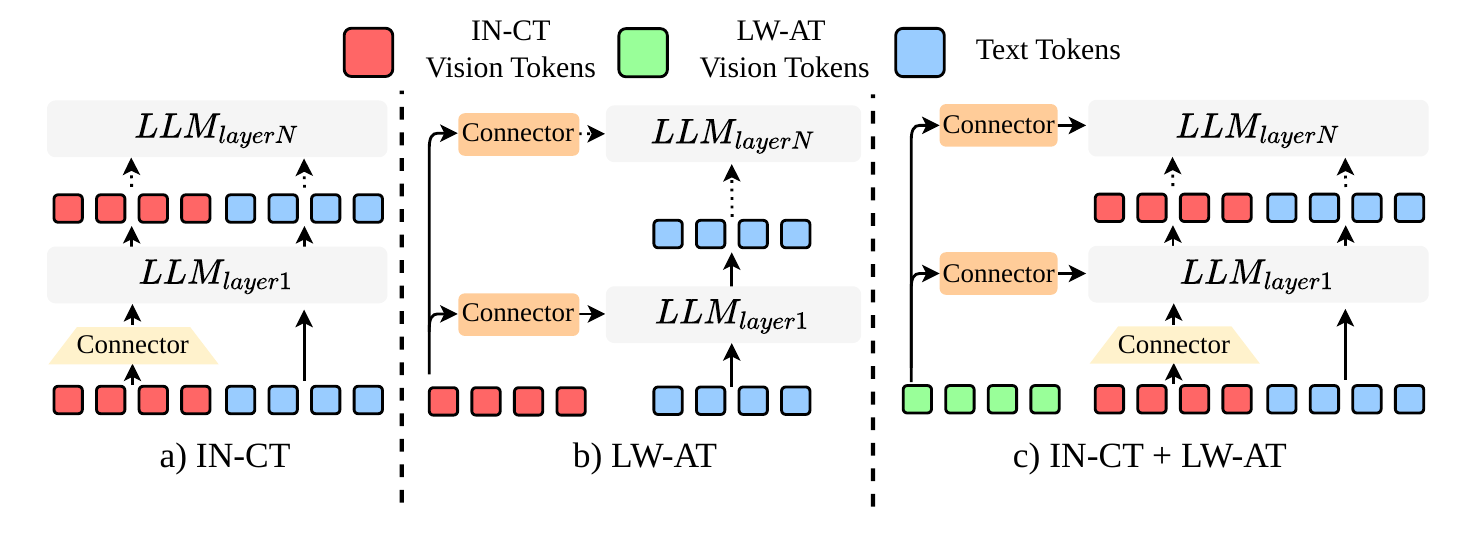}
  \caption{Architecture overview of the hybrid IN-CT + LW-AT model compared to standalone IN-CT and LW-AT. IN-CT vision tokens are concatenated at the input, while LW-AT vision tokens are injected into the keys and values at every layer.}
  \label{app_fig:appendix_hybrid_arch}
\end{figure}

We provide full implementation details for the hybrid model introduced in the main paper. The hybrid model (IN-CT + LW-AT) applies both integration strategies simultaneously. As illustrated in Figure~\ref{app_fig:appendix_hybrid_arch}, IN-CT vision tokens are concatenated with text tokens at the input following the standard in-context injection pipeline, while LW-AT vision tokens are additionally injected into the keys and values of each LLM layer.

\begin{figure}[!h]
  \centering
  \includegraphics[width=\linewidth]{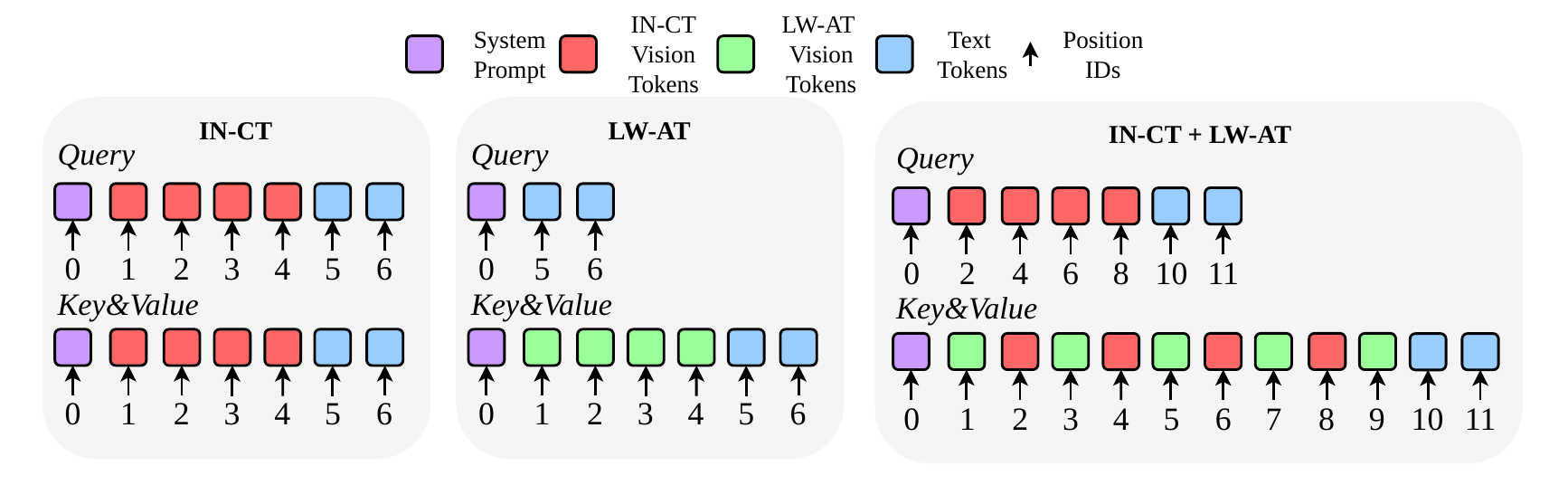}
  \caption{Token sequence and position ID assignment for IN-CT, LW-AT, and the hybrid IN-CT + LW-AT model. Each paradigm preserves its original position ID scheme independently within the hybrid architecture.}
  \label{app_fig:appendix_vlm_pos}
\end{figure}

The two sets of visual tokens are processed by separate connectors but share the same vision encoder features. Figure~\ref{app_fig:appendix_vlm_pos} illustrates the resulting token sequence and position ID assignment. IN-CT vision tokens retain their original position IDs within the input sequence, while LW-AT vision tokens are assigned position IDs in the key-value sequence following the same scheme as the standalone LW-AT model, preserving the relative positional structure of each paradigm independently. 

\section{Training Details}
\label{app:a2_training_details}

\subsection{Input Image Processing}\label{app:a2_image_processing}
For training with the 700K and text-centric-filtered 500K mixtures, which consist entirely of single images, we apply standard image resizing to $384 \times 384$. For training with the LLaVA-OV mixture, which combines high-resolution single images, multi-image, and video data, we follow the AnyRes input processing pipeline from~\citet{li2024llava}. Specifically, high-resolution single images are processed using the same adaptive tiling strategy, while multi-image and video inputs follow the same frame sampling and sequence construction as in the original LLaVA-OV setup. This input processing pipeline is applied identically across all three integration paradigms (IN-CT, LW-AT, and LW-GC), ensuring that any observed performance differences are attributable solely to the integration mechanism rather than to differences in visual input handling.

\subsection{Hyperparameters and Computing Infrastructure}\label{app:a2_hparams}
All models are trained using the same two-stage pipeline and hyperparameter configuration, summarised in Table~\ref{tab:training_settings}. All paradigms share identical settings across both stages, ensuring that any observed differences are attributable solely to the integration mechanism. All experiments are conducted on NVIDIA H200 GPUs, with the per-stage GPU allocation listed in the same table.

\begin{table*}[!h]
\centering
\begin{tabular}{lcc}
\toprule
\textbf{Configuration} & \textbf{Stage 1: Pretraining} & \textbf{Stage 2: Supervised Fine-Tuning} \\
\midrule
\multicolumn{3}{l}{\textit{Trainable modules}} \\
\midrule
Vision encoder (SigLIP2-So400m) & Frozen      & Trainable \\
MLP projector                & Trainable   & Trainable \\
LLM (Qwen2.5 / Llama-3.2)    & Frozen      & Trainable \\
\midrule
\multicolumn{3}{l}{\textit{Training configuration}} \\
\midrule
Dataset                      & LLaVA-Pretrain  & LLaVA-NeXT / LLaVA-OV \\
Global batch size            & 256             & 256 \\
Learning rate (projector)    & $1\times10^{-3}$ & $1\times10^{-5}$ \\
Learning rate (VE)           & --              & $2\times10^{-6}$ \\
Learning rate (LLM)          & --              & $1\times10^{-5}$ \\
Optimizer                    & AdamW           & AdamW \\
Weight decay                 & 0.0             & 0.0 \\
LR schedule                  & Cosine decay    & Cosine decay \\
Warmup ratio                 & 0.03            & 0.03 \\
Epochs                       & 1               & 1 \\
Distributed training         & DeepSpeed ZeRO-2 & DeepSpeed ZeRO-2 \\
\multirow{2}{*}{Image resolution} & \multirow{2}{*}{$384\times384$} & $384\times384$ (LLaVA-NeXT) \\
                             &                 & AnyRes (LLaVA-OV) \\
Precision                    & bfloat16        & bfloat16 \\
GPU                          & NVIDIA H200     & NVIDIA H200 \\
GPU count                    & 4 (1 node)      & 16 (4 nodes) \\
\bottomrule
\end{tabular}
\caption{Training hyperparameters, configurations, and compute allocation for the two-stage pipeline. In Stage~1, only the connector is trained to align visual features. In Stage~2, the entire architecture (vision encoder, projector, and LLM) undergoes full-parameter instruction tuning. For Stage~2, we experiment with two SFT mixtures: LLaVA-NeXT (700K) and LLaVA-OV (1.6M).}
\label{tab:training_settings}
\end{table*}

\subsection{Training Dataset Details}\label{app:a2_datasets}
For Stage~1 pretraining, we use LLaVA-Pretrain (BLIP-558K)~\citep{liu2023visual}, a standard image-caption dataset used to align the visual connector with the LLM. For Stage~2 supervised fine-tuning, we experiment with three mixtures. LLaVA-NeXT SFT (700K)~\citep{liu2024llavanext} is a general visual instruction tuning mixture covering diverse question answering, reasoning, and conversation tasks. The Text-Centric-Filtered (500K) mixture is derived from LLaVA-NeXT SFT by removing text-centric samples, including pure OCR datasets as well as samples containing substantial situated text such as OCR-VQA~\citep{mishra2019ocr}, DocVQA~\citep{mathew2021docvqa}, AI2D~\citep{kembhavi2016diagram}, ChartQA~\citep{masry2022chartqa}, DVQA~\citep{kafle2018dvqa}, and Synthdog~\citep{kim2022ocr}, resulting in approximately 500K samples after filtering. LLaVA-OV (1.6M)~\citep{li2024llava} is a large-scale mixture covering single-image, multi-image, and video data.

\section{Centered Kernel Alignment (CKA)}\label{app:cka}

We employ Centered Kernel Alignment (CKA)~\citep{kornblith2019similarity} to measure how token representations evolve across layers. Here we provide the formal definition.

\paragraph{Formal definition.}
Given two representation matrices $X \in \mathbb{R}^{T \times D}$ and $Y \in \mathbb{R}^{T \times D}$, where $T$ is the number of tokens and $D$ is the hidden dimension, we define their kernel matrices as $K = XX^\top$ and $L = YY^\top$ using a linear kernel. CKA is then computed via the Hilbert-Schmidt Independence Criterion (HSIC):
\begin{equation}
    \text{CKA}(K, L) = \frac{\text{HSIC}(K, L)}{\sqrt{\text{HSIC}(K, K) \cdot \text{HSIC}(L, L)}}
\end{equation}
where HSIC measures the statistical dependence between the two feature spaces. For centered kernel matrices $\tilde{K} = HKH$ and $\tilde{L} = HLH$, with $H = I_T - \frac{1}{T}\mathbf{1}\mathbf{1}^\top$ being the centering matrix, the empirical HSIC is given by:
\begin{equation}
    \text{HSIC}(K, L) = \frac{1}{(T-1)^2} \operatorname{tr}(\tilde{K}\tilde{L})
\end{equation}

We apply CKA at the token level to measure how the internal structure of visual or text token representations shifts across layers. For a batch of $N$ samples, let $H^{(\ell)}_i \in \mathbb{R}^{T \times D}$ denote the hidden states of sample $i$ at layer $\ell$. We treat the $T$ tokens as data points and compute:
\begin{equation}
    \text{CKA}_{\text{token}}(\ell, \ell') = \frac{1}{N} \sum_{i=1}^{N} \text{CKA}(H^{(\ell)}_i, H^{(\ell')}_i)
\end{equation}
A high $\text{CKA}_{\text{token}}(\ell, \ell')$ indicates that the pairwise relationships among tokens are largely preserved between layers $\ell$ and $\ell'$, whereas a low value signals a structural reorganisation of the token representations.

\section{Additional Attention Mass Results}\label{app:attn_mass}
Section~\ref{sec:finding4_attention} of the main paper reports attention mass for IN-CT and LW-AT on one general and one text-centric benchmark. Here we extend the same analysis to additional benchmarks in Figure~\ref{app_fig:SUP_attention_mass_supplementary} below. The task-dependent pattern holds throughout. Attention to visual tokens stays low across all layers on non-OCR benchmarks but remains substantially higher through the middle and deep layers on text-centric ones. The two paradigms allocate attention almost identically, reinforcing that the performance gap between them reflects representation quality rather than how much attention the visual tokens receive.

\begin{figure*}[h]
  \centering
  \includegraphics[width=\linewidth]{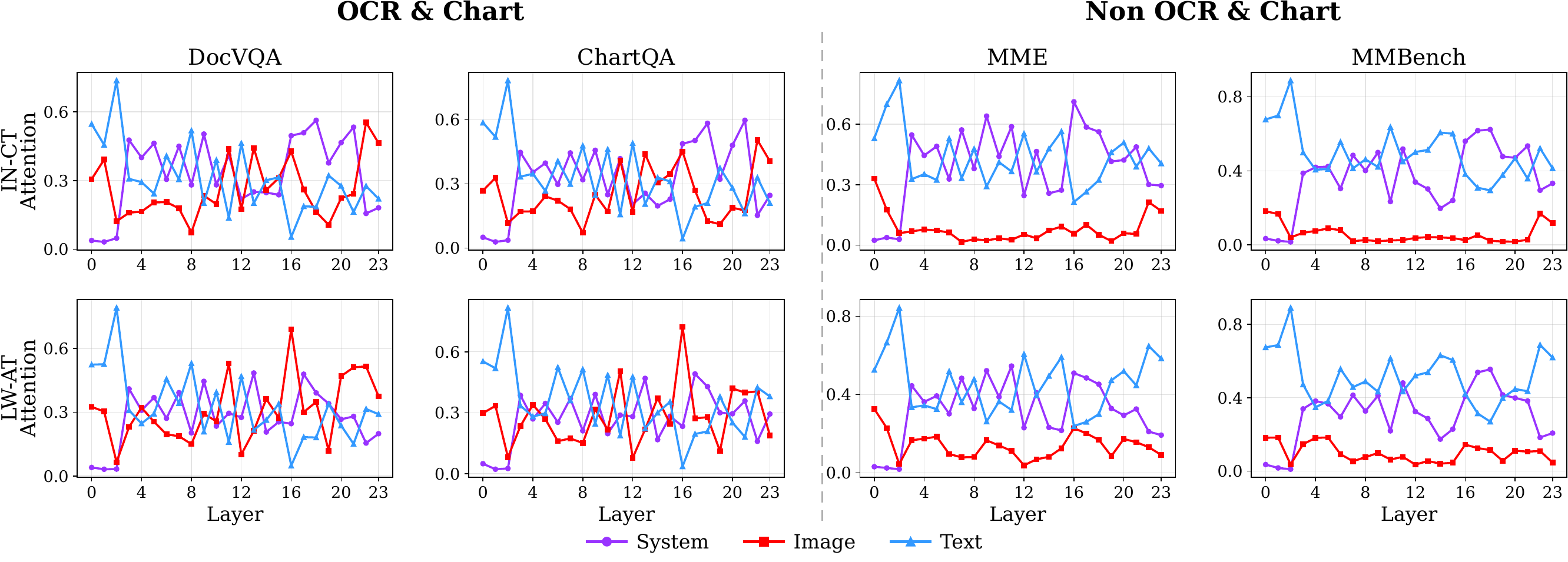}
\caption{Attention mass allocated to visual tokens during generation across layers on additional benchmarks for both IN-CT and LW-AT. Consistent with the main paper, attention patterns are task-dependent, with text-centric tasks maintaining substantially higher attention to visual tokens through the deeper layers than non-OCR tasks.}
  \label{app_fig:SUP_attention_mass_supplementary}
\end{figure*}

\section{Additional PCA Visualisations}\label{app:pca}
Figure~\ref{fig:Fig4_fft_analysis_and_fft_analysis} in the main paper shows PCA projections at five representative layers. Figures~\ref{app_fig:SUP_pca_all_layers_in_ct} to~\ref{app_fig:SUP_pca_all_layers_lw_gc} below show every layer for each paradigm. The convergence in IN-CT is visible across the full depth rather than only at the layers selected for display, and no such convergence appears at any depth in either layer-wise variant.

\begin{figure*}[ht]
  \centering
  \includegraphics[width=0.95\linewidth]{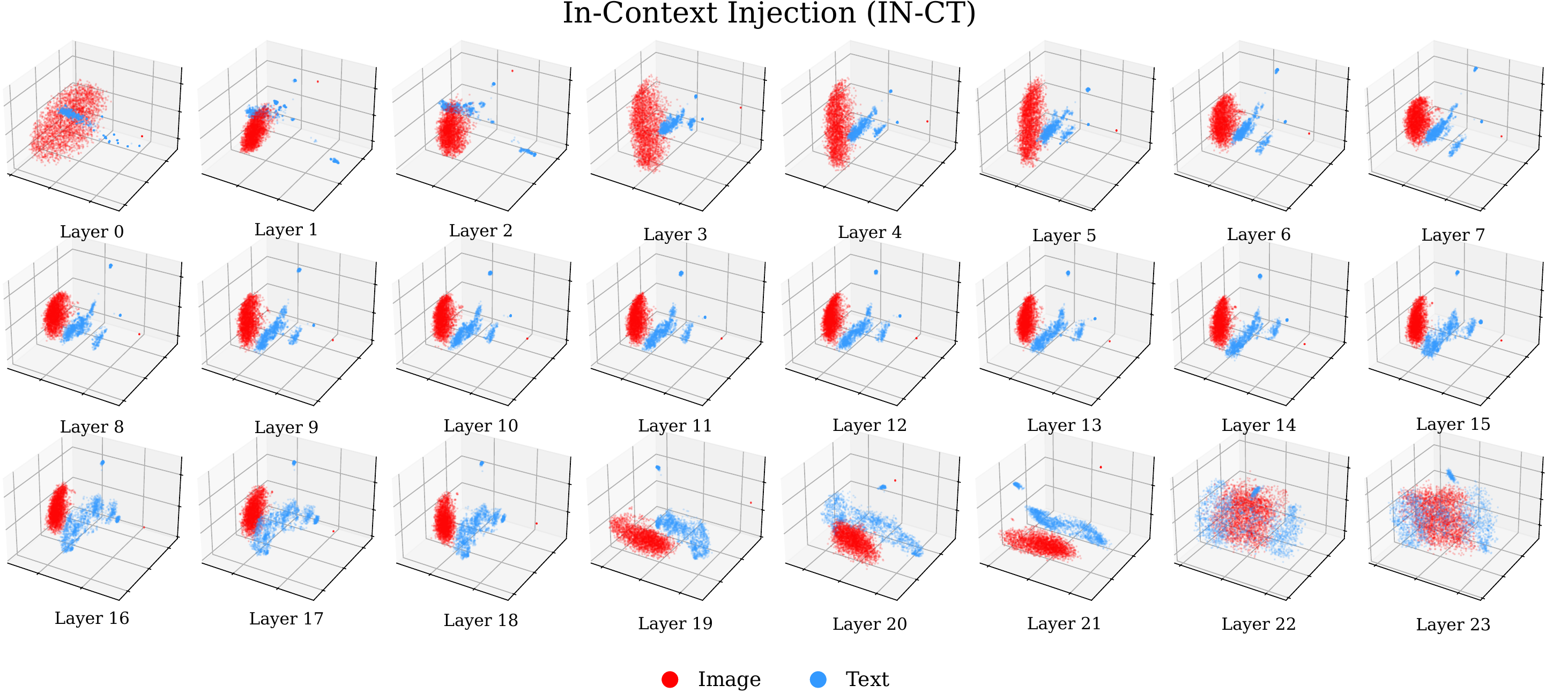}
\caption{3D PCA projections of image (red) and text (blue) token representations at every layer for IN-CT. Visual tokens progressively converge toward the text token subspace, with clear merging in the final layers.}
  \label{app_fig:SUP_pca_all_layers_in_ct}
\end{figure*}

\begin{figure*}[ht]
  \centering
  \includegraphics[width=0.95\linewidth]{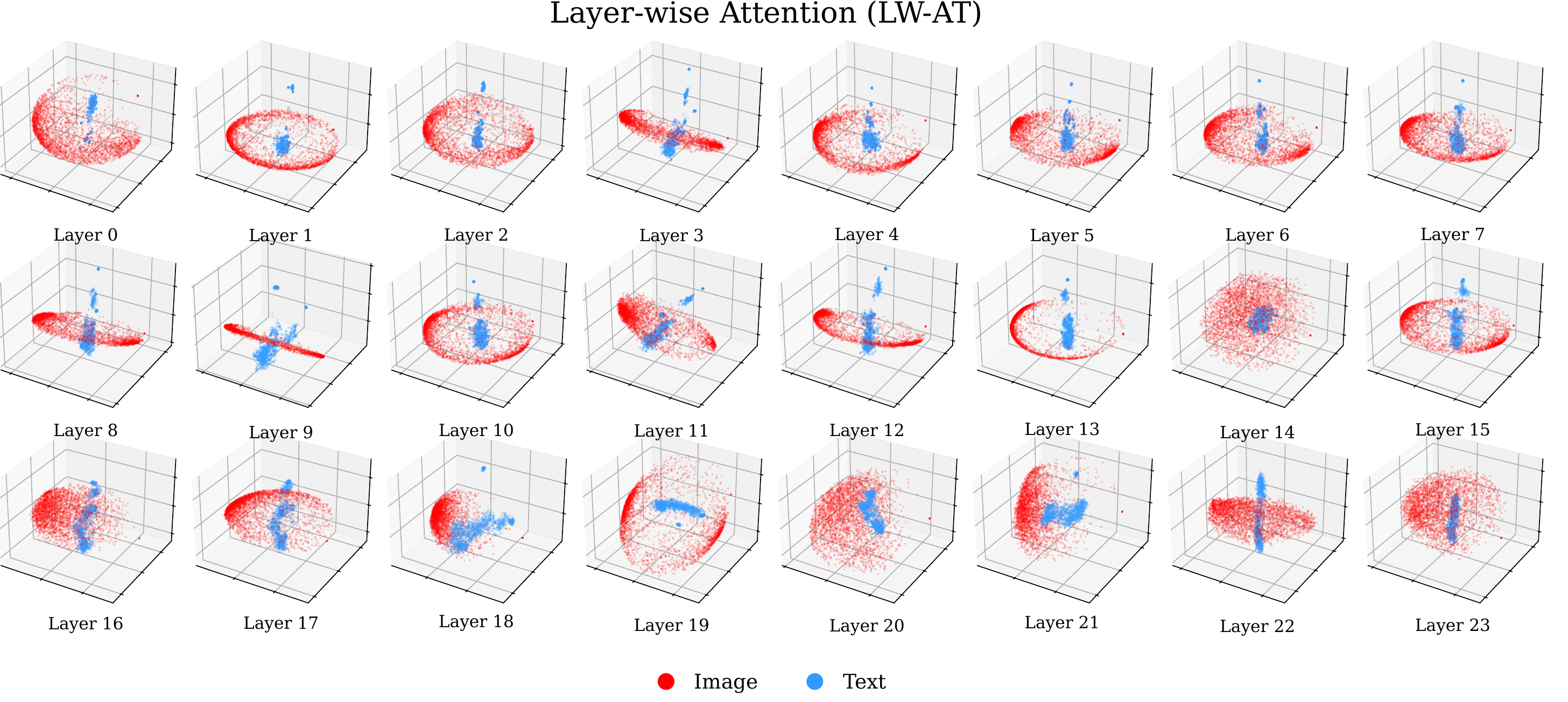}
\caption{3D PCA projections of image (red) and text (blue) token representations at every layer for LW-AT. Visual representations remain separated from the text subspace throughout all layers.}
  \label{app_fig:SUP_pca_all_layers_lw_at}
\end{figure*}

\begin{figure*}[ht]
  \centering
  \includegraphics[width=0.95\linewidth]{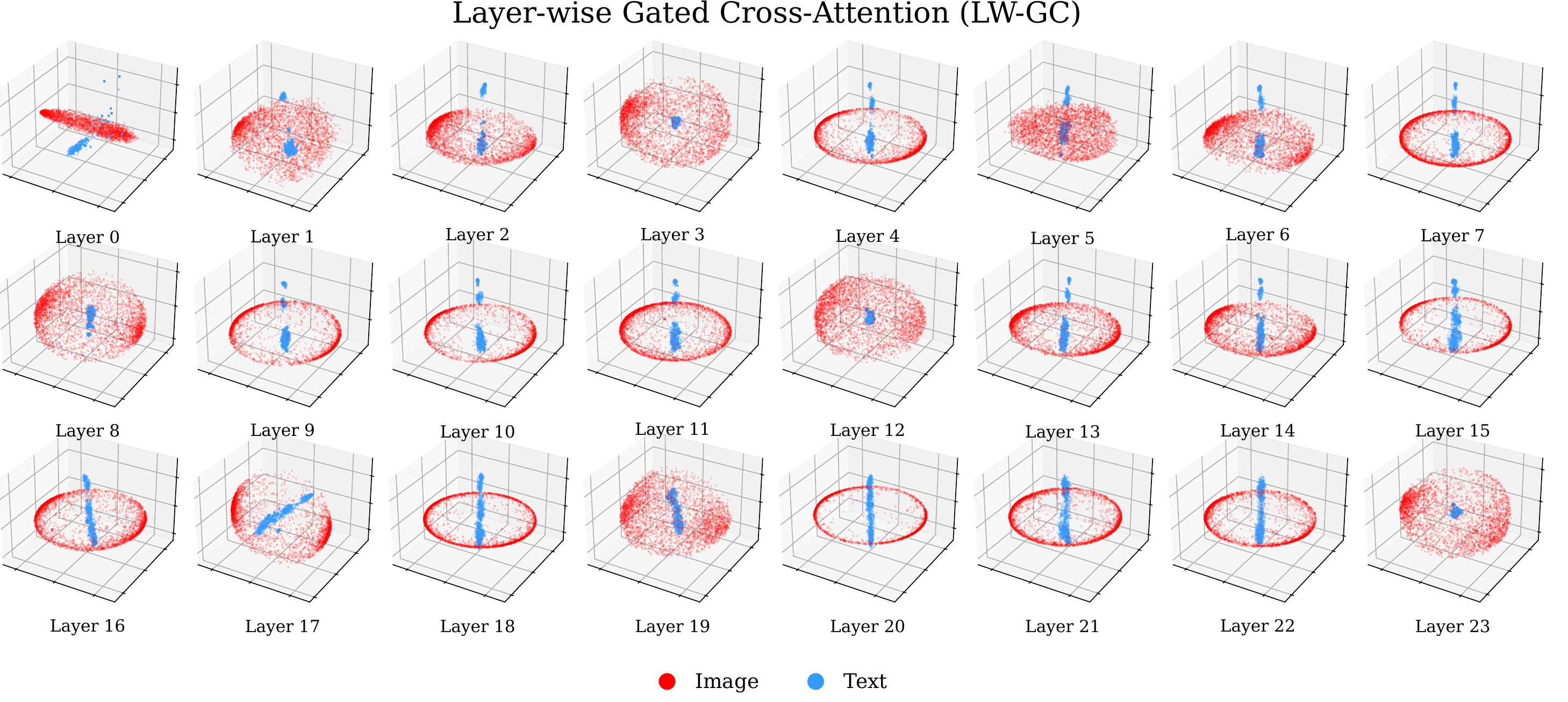}
\caption{3D PCA projections of image (red) and text (blue) token representations at every layer for LW-GC. As with LW-AT, visual and text representations occupy distinct subspaces throughout the network.}
  \label{app_fig:SUP_pca_all_layers_lw_gc}
\end{figure*}

\end{document}